\documentclass[10pt,twocolumn,letterpaper]{article}

\usepackage{multirow}
\usepackage{booktabs}
\usepackage{colortbl}
\usepackage{makecell}
\usepackage{tikz}

\usepackage{pifont}       

\definecolor{forestgreen}{RGB}{34,139,34}
\newcommand{\cmark}{\ding{51}}
\newcommand{\xmark}{\ding{55}}

\usepackage{cvpr}              

\definecolor{cvprblue}{rgb}{0.21,0.49,0.74}
\usepackage[pagebackref,breaklinks,colorlinks,citecolor=cvprblue]{hyperref}

\def\paperID{1702} 
\def\confName{CVPR}
\def\confYear{2024}

\title{Language-driven All-in-one Adverse Weather Removal}

\author{Hao Yang$^{1}$, \quad Liyuan Pan$^{1}$, \quad Yan Yang$^{2}$, \quad and \quad Wei Liang$^{1}$ \\
$^{1}$Beijing Institute of Technology \quad $^{2}$ The Australian National University \\
{\tt\small \{hao.yang, liyuan.pan, liangwei\}@bit.edu.cn, \quad \{yan.yang\}@anu.edu.au}
}

\begin{document}
\maketitle
\begin{abstract}
All-in-one (AiO) frameworks restore various adverse weather degradations with a single set of networks jointly. 
To handle various weather conditions, an AiO framework is expected to adaptively learn weather-specific knowledge for different degradations and shared knowledge for common patterns. 
However, existing methods: 1) {rely on} extra supervision signals, which are usually unknown in real-world applications;  2) {employ fixed network structures}, which restrict the diversity of weather-specific knowledge.
In this paper, we propose a Language-driven Restoration framework (LDR) to alleviate the aforementioned issues. 
First, we leverage the power of pre-trained vision-language (PVL) models to enrich the diversity of weather-specific knowledge by reasoning about the occurrence, type, and severity of degradation, generating description-based degradation priors.
Then, with the guidance of degradation prior, we sparsely select restoration experts from a candidate list dynamically based on a Mixture-of-Experts (MoE) structure. This enables us to adaptively learn the weather-specific and shared knowledge to handle various weather conditions (\eg, unknown or mixed weather). 
Experiments on extensive restoration scenarios show our superior performance (see Fig.~\ref{fig:radar}). The source code will be made available.
\end{abstract}

\section{Introduction}

Imaging under adverse weather conditions leads to unpleasant image degradation, {posing challenges for vision-based systems like self-driving cars \cite{hu2023planning,xiong2023neural} and outdoor surveillance systems \cite{doshi2022multi,escalera2023surveillance}}, that require 24/7 service regardless of rain \cite{chen2023learning,fu2023continual,liang2022drt}, haze \cite{song2023vision,hoang2023transer,qin2020ffa}, and snow \cite{chen2021all,liu2018desnownet,mokayed2023nordic}. To meet the safety demand, compared to tackling each type of weather condition with independent models, the joint task process, \ie, all-in-one framework, has a broader application scenario \cite{park2023all,li2022all}.

\begin{figure}[!t]
    \centering
    \includegraphics[width=0.9\linewidth]{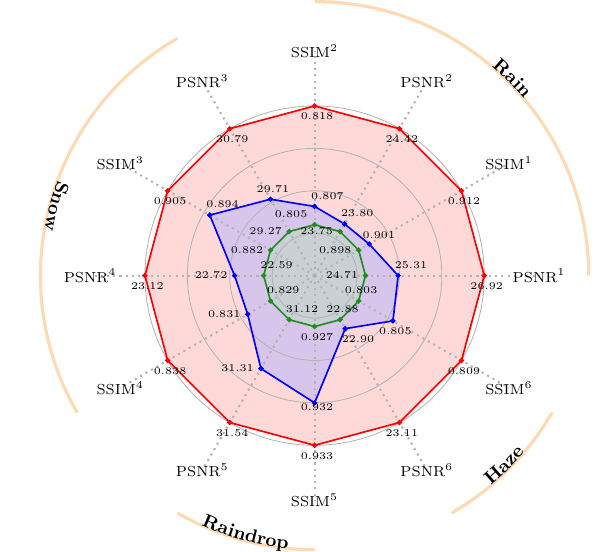}
    \vspace{-0.5em}
    \caption{\small \it 
    Score (PSNR and SSIM) comparisons. We compare our model (\textcolor{red}{red}) with the top 2 (\textcolor{blue}{blue} and \textcolor{forestgreen}{green}) baselines on benchmark datasets  with various weather scenarios. Superscripts besides evaluation metrics are used to differentiate benchmark datasets.  
    }
    \label{fig:radar}
    \vspace{-0.5em}
\end{figure}

Formally, degraded images can be modeled as masked {additive combination} of clean images and degradation residuals\cite{park2023all}. 
{Consequently}, several works \cite{li2020all, chen2022learning, valanarasu2022transweather} use a single network for all degradation types. Though shared knowledge is learned for restoration, they neglect that different degradations still hold different mathematical formulations, \eg, transmission map produced by scattering effect in haze model \cite{song2023vision} is {unnecessary} for raindrop model \cite{qian2018attentive}.


Hence, several works \cite{li2022all,park2023all, zhu2023learning,patil2023multi,ye2023adverse} use different sub-networks for weather-specific knowledge learning. 
However, auxiliary supervisions are required to assign the sub-networks, \eg, degradation types \cite{park2023all} or depth maps \cite{zhu2023learning}. Furthermore, the fixed sub-network architecture of existing methods restricts the diversity of learned weather-specific knowledge and their ability to handle images with various weather conditions \cite{zhang2023infwide, wan2022image}, such as images degraded by weather severity or conditions that have not been encountered before \cite{qian2018attentive,chen2021pre}, or mixed weather conditions like snow with haze in real-world scenarios.

\begin{figure*}[!t]
    \centering
    \begin{minipage}{.65\linewidth}
    \begin{tikzpicture}
    \node[anchor=south west,inner sep=0] (image) at (0,0) {\includegraphics[width=\linewidth]{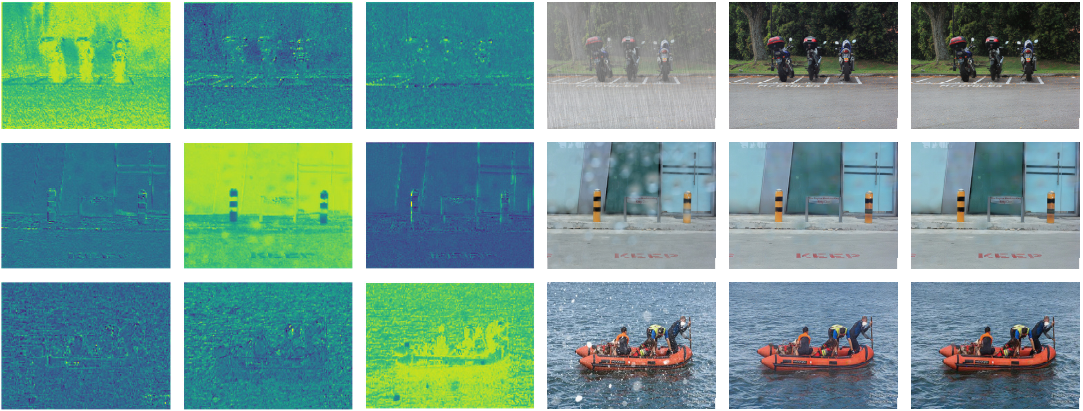}};
    \begin{scope}[x={(image.south east)},y={(image.north west)}] 
        \draw (0.07,  -0.07) node {\small (a) 7-th};
        \draw (0.24,  -0.07) node {\small (b) 10-th};
        \draw (0.41,  -0.07) node {\small (c) 54-th};
        \draw (0.58,  -0.07) node {\small (d) Input};
        \draw (0.75,  -0.07) node {\small (e) Transweather};
        \draw (0.92,  -0.07) node {\small (f) Zero-out};
        \draw (1.15,  -0.07) node {\small (g) Distribution};
    \end{scope}
    \end{tikzpicture}
    \vspace{-2.35em}
    \end{minipage}
    \begin{minipage}{.23\linewidth}
    \begin{tikzpicture}
    \node[anchor=south west,inner sep=0] (image) at (0,0) {\includegraphics[width=\linewidth]{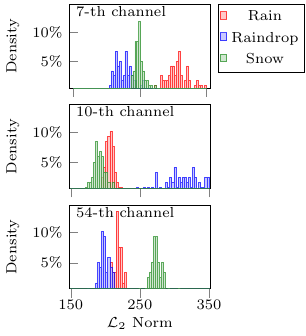}};
    \begin{scope}[x={(image.south east)},y={(image.north west)}] 
    \end{scope}
    \end{tikzpicture}
    \end{minipage}
    \vspace{-0.1em}
    \caption{\small \it  Ineffective computations of Transweather \cite{valanarasu2022transweather}. We find the activity of some parameters in Transweather \cite{valanarasu2022transweather} are degradation dependent, and zero-outing computations for those inactivate parameters barely affect image restoration. (a) 7-th channel, (b) 10-th channel, and (c) 54-th channel of feature maps from Transweather, the brighter, the larger activation value, are activated for degradation of rain, raindrop, and snow. We show the (d) degraded images, (e) restorations from Transweather, and (f) restorations by zero-outing inactive channels. (g) From top to bottom, distribution of $\mathcal{L}_{2}$ norm for the three channels across all images in the All-weather dataset \cite{li2020all}. 
    }  
    \label{img:channel_vis}
    \vspace{-0.5em}
\end{figure*}
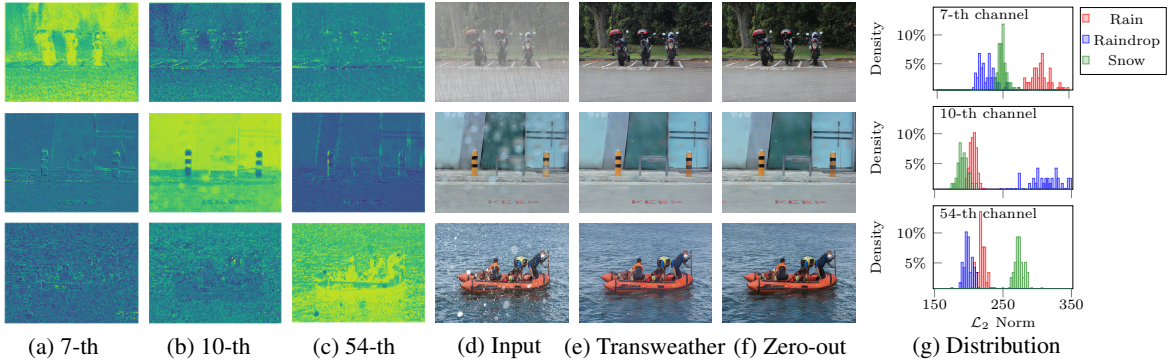

In this paper, we question -- can we restore the image degraded by various weather conditions by adaptive {learning of} diverse weather-specific knowledge and shared knowledge, without requiring ground truth weather types and severity data? We answer it by our {\it \underline{l}anguage-\underline{d}riven all-in-one \underline{r}estoration} (LDR) framework in two aspects.


{\bf 1)} The knowledge within the feature space of a pre-trained vision-language (PVL) model can benefit various tasks, while its potential in our task is still under exploration. A straightforward way is to use the PVL model as the image degradation classifier. In contrast, we go one step further by reasoning diverse weather-specific knowledge from the feature space of the PVL model beyond the type of weather conditions.



We start by formulating a question prompt to query the occurrence, type, and severity of degradation in a degraded image. The obtained degradation prior describes {\emph{what, where, and severity}} of degradations in high-level {textual} semantics. Then, we translate the high-level degradation prior to a 2D degradation map by aligning the prior with the degraded image. This degradation map provides a pixel-wise representation of the diverse knowledge of image degradation from the PVL model.


{\bf 2)} We then unleash the potential for various weather removal with the guidance of degradation maps. Observing model parameters are weather-specific, \eg, the rain-related parameters are usually inactive for unrelated degradations, and zero-outing the computations of unrelated parameters also barely affects the restoration quality. {This observation, illustrated in \cref{img:channel_vis}, inspired us to bypass computations for parameters unrelated to the specific weather type and severity during image restoration.} With the assistance of MoE structure \cite{masoudnia2014mixture,shazeer2017outrageously}, our LDR framework selects experts dynamically for restoration, therefore, ensuring adaptive learning of weather-specific knowledge which is not limited to a fixed network architecture.

Specifically, we maintain a candidate list of restoration {experts} and {utilize} the degradation map to sparsely select the most related restoration experts for each degradation. By applying the selected expert pixel-wisely to restore weather-specific features, we create flexible and degradation-adaptive expert combinations/model architectures for restoration.


Though we can have a preliminary restoration from experts restored feature, considering image regions with similar values from the degradation map tend to benefit restoration features of each other, we re-use the degradation map to aggregate the restoration features, and improve the locality of the obtained restoration by a simple convolutional feedforward network. Our main contributions are:
%
%
\begin{itemize}
    \setlength\itemsep{-0.1em}
    \item We present an LDR framework to adaptively remove various adverse weather conditions in an all-in-one solution;
    \item We propose a degradation map measurement module for extracting diverse weather-specific knowledge from a pre-trained vision-language model;
    \item We propose a Top-K expert restoration module, sparsely and adaptively computing pixel-wise restoration features. 
\end{itemize}

The overall comparison in Fig.~\ref{fig:radar} shows the superiority of our framework for handling various degradations compared to state-of-the-art methods. 


\begin{figure*}[!t]
    \centering
    \includegraphics{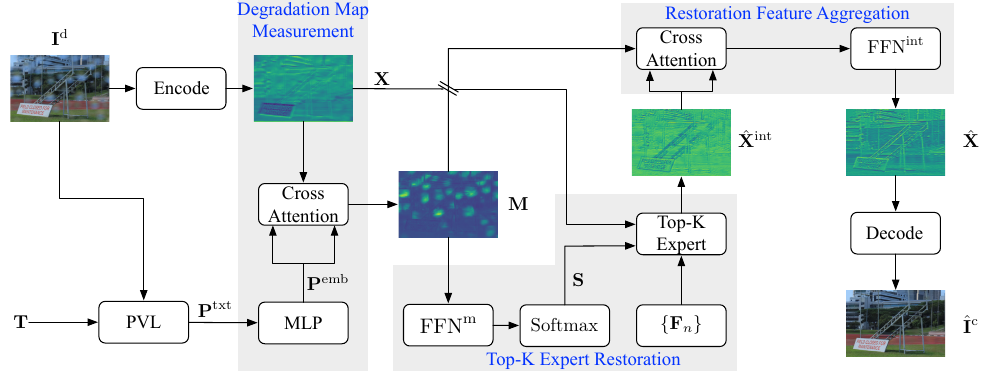}
    \vspace{-1.11em}
    \caption{\small \it The pipeline of our method. Given the input degraded image  $\mathbf{I}^{\text{d}}$, we aim to recover the clean image $\hat{\mathbf{I}}^{\text{c}}$.  To tackle various adverse weather conditions, we format a question prompt $\mathbf{T}$, and query a pre-trained vision-language (PVL) model with $\mathbf{I}^{\text{d}}$ and $\mathbf{T}$, to estimate the degradation type and severity. The generated descriptions $\mathbf{P}^{\text{txt}}$ are transformed by a multilayer perceptron (MLP) to get the degradation prior  $\mathbf{P}^{\text{emb}}$. Meanwhile, we extract a feature map $\mathbf{X}$ from the input degraded image  $\mathbf{I}^{\text{d}}$. A degradation map $\mathbf{M}$ is computed by cross-attending $\mathbf{X}$ with the prior $\mathbf{P}^{\text{emb}}$, describing pixel-wise degradation pattern. We maintain a trainable candidate list of convolution filters, $\{\mathbf{F}_{n}|{n=1,...,N}\}$, and denote them as experts. We first parse the degradation map $\mathbf{M}$ into a score map $\mathbf{S}$ via a feedforward network $\text{FFN}^{\text{m}}$ and a softmax layer, and then use $\mathbf{S}$ to find the best $K$ experts describing the degradation of $\mathbf{X}$. Best experts are convolved with $\mathbf{X}$ to 
    generate an intermediate feature map $\hat{\mathbf{X}}^\text{int}$. Finally, a cross-attention layer is used to aggregate $\hat{\mathbf{X}}^\text{int}$ with the guidance of the degradation map $\mathbf{M}$. We improve the feature locality with a feedforward network $\text{FFN}^{\text{int}}$, and the output $\hat{\mathbf{X}}$ is decoded to the clean image $\hat{\mathbf{I}}^\text{c}$.  
    }
    \label{fig:arch}
    \vspace{-0.5em}
\end{figure*}
 
\section{Related Work}


\paragraph{Adverse Weather Restoration.} The field of adverse weather restoration {encompasses} two distinct approaches: task-specific methods \cite{zamir2021multi,chen2021hinet,chen2022simple,zamir2022restormer,li2023efficientgrl} train the model independently for each individual degradation and all-in-one frameworks \cite{li2020all,chen2022learning,li2022all,valanarasu2022transweather,park2023all,zhu2023learning,wang2023smartassign,zhang2023ingredient} develop a single unified model capable of handling various adverse weather degradations. A {prevalent strategy} for the all-in-one framework is to build upon task-specific models by employing multi-task learning techniques \cite{zhu2023learning,wang2023smartassign,zhang2023ingredient} or knowledge distillation \cite{chen2022learning} that consolidates multiple task-specific networks into a single network. Nevertheless, these methods typically rely on fixed computations, and the computation effectiveness is discounted as model parameters are separately and degradation-dependently activated. 
%
%

Conversely, several methods \cite{li2022all,park2023all,zhu2023learning} have been developed to overcome these limitations by focusing on learning weather-specific knowledge. These methods utilize indicators like degradation types \cite{park2023all} or depth maps \cite{zhu2023learning} during training to categorize and direct images to appropriate sub-networks for restoration. Though AirNet \cite{li2022all} learns weather-specific knowledge with contrastive learning. Its fixed sub-network designs can only restore images with certain degradation types, and does not generalize to unseen weather degradations, \eg, rain mixed with haze. Furthermore, these mentioned works overlook the fact that varying levels of degradation severity also warrant adaptively tailored computational processes, for a specific degradation type.
%
{Diverging from previous methods, we harness PVL models to reason both the type and severity of degradations, enabling adaptive restoration with dynamic sub-networks informed by this rich, weather-specific knowledge.}

\vspace{-5.5mm}
\paragraph{Sparse Mixture of Expert.} The concept of expert models \cite{masoudnia2014mixture} is defined as a subset of model parameters/computation, where each expert is specialized in handling distinct aspects of the input data. The sparse mixture of experts usually employs a routing mechanism \cite{shazeer2017outrageously} to dynamically and adaptively forward input to a subset of these experts, bypassing unnecessary and irrelevant computations. Given fixed computation costs, due to the sparsity, this framework can {readily} scale the number of experts to improve the model capability that has been widely verified in the domain of natural language processing \cite{shazeer2017outrageously,du2022glam,zuo2022moebert} and computer vision \cite{ng2023botbuster,enzweiler2011multilevel,li2023meid,chen2022towards}. We study a prior derived from a PVL model to dynamically select experts, and adaptively apply expert to restore degraded images. 


\vspace{-5mm}
\paragraph{Vision-language Model.} With the release of ChatGPT, remarkable reasoning abilities of large language model (LLM) \cite{ordonez2011im2text,radford2021learning,lai2023lisa,liu2023visual,zhu2023minigpt,li2022blip} have been shown. It motivates the vision-language community to transfer the reasoning abilities to visual data. A common pipeline is to project an image to a joint space of LLM \cite{li2023blip,liu2023visual, liang2023iterative}, and feed the projected image alongside {user-provided text} for conditional text generation. This paper proposes to leverage the PVL (\eg, \cite{lai2023lisa}) to generate prior of an adverse weather degraded image for adaptive image restoration.



\section{Method}

\paragraph{Overview.} Given an {input} image $\mathbf{I}^{\text{d}}$ degraded by adverse weather, we aim to recover the {corresponding} clean image $\hat{\mathbf{I}}^{\text{c}}$. 
Our method adaptively recovers $\hat{\mathbf{I}}^{\text{c}}$ with degradation prior $\mathbf{P}^{\text{emb}}$ {obtained} from a pre-trained vision-language (PVL) model.  The overall architecture is shown in Fig.~\ref{fig:arch}.  

{To derive} degradation prior $\mathbf{P}^{\text{emb}}$, 
{we query a PVL model with a question prompt $\mathbf{T}$ to reason about the occurrence, type, and severity of degradation in the input image $\mathbf{I}^{\text{d}}$, outputting $\mathbf{P}^{\text{txt}}$.}
Then a mapping network projects $\mathbf{P}^{\text{txt}}$ to $\mathbf{P}^{\text{emb}}$. Meanwhile, we {encode} $\mathbf{I}^{\text{d}}$ to $\mathbf{X}$ {by using an encoder}.

{We adaptively restore $\mathbf{X}$ into $\hat{\mathbf{X}}$ in the embedding space, and decode $\hat{\mathbf{X}}$ to a restored image $\hat{\mathbf{I}}^{\text{c}}$, with three steps:}
i) degradation map measurement,  measuring degradation for each pixel of $\mathbf{I}^{\text{d}}$ by using the degradation prior $\mathbf{P}^{\text{emb}}$;
ii) top-K expert restoration, selecting expert/parameters from a {trainable candidate list of convolution filters} according to the degradation map $\mathbf{M}$, and convolving with ${\mathbf{X}}$ to get intermediate restoration features $\hat{\mathbf{X}}^{\text{int}}$;
iii) restoration feature aggregation, deriving $\hat{\mathbf{X}}$ by pixel-wisely aggregating $\hat{\mathbf{X}}^{\text{int}}$ with respect to the degradation map $\mathbf{M}$, and improving the feature locality with a feedforward network. 
Finally, $\hat{\mathbf{X}}$ is decoded to get the restored image $\hat{\mathbf{I}}^\text{c}$.

\begin{figure}[!t]
    \centering
    \begin{tikzpicture}
    \node[anchor=south west,inner sep=0] (image) at (0,0) {\includegraphics[width=0.95 \linewidth]{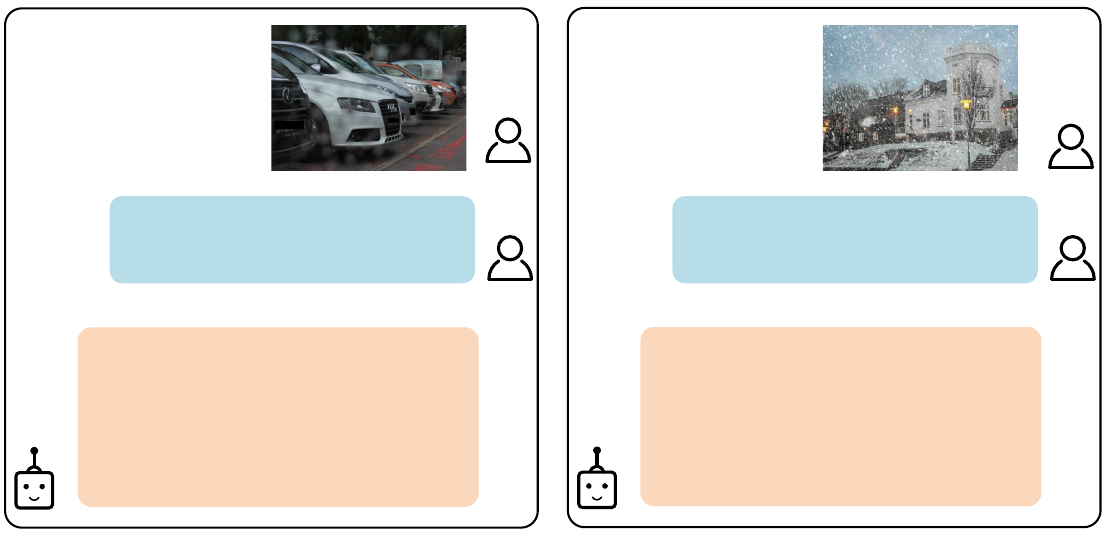}};
    \begin{scope}[x={(image.south east)},y={(image.north west)}]

\draw (0.255, 0.52) node[align=left] {
    \tiny Please describe about the weather \\[-1.3ex]
    \tiny  in the picture. \\[-1.3ex]
};
    
\draw (0.25, 0.18) node[align=left] {
    \tiny The weather in the picture is rainy, as  \\[-1.3ex]
    \tiny  evidenced by the presence of raindrop \\[-1.3ex]
    \tiny  on the cars and the overall atmosphere.  \\[-1.3ex]
    \tiny  \\[-1.4ex]
};

\draw (0.77, 0.53) node[align=left] {
    \tiny Please describe the type of weather,\\[-1.5ex]
    \tiny  intensity, and obscured areas in the \\[-1.5ex]
     \tiny  picture. \\[-1.5ex]
};

\draw (0.76, 0.21) node[align=left] {
    \tiny The weather in the image is snowy, the \\[-1.3ex]
    \tiny intensity of the snowfall is high. The   \\[-1.3ex]
    \tiny   area covered by the snow includes a    \\[-1.3ex]
    \tiny   residential neighborhood, with houses   \\[-1.3ex]
    \tiny   in the scene are obscured by snowflakes.  \\[-1.3ex]
};


    \end{scope}
    \end{tikzpicture}
    \vspace{-1em}
    \caption{\small \it
    Examples text descriptions from LISA \cite{lai2023lisa} for reasoning weather degraded images. 
    }
    \label{img:vqa}
    \vspace{-0.5em}
\end{figure}
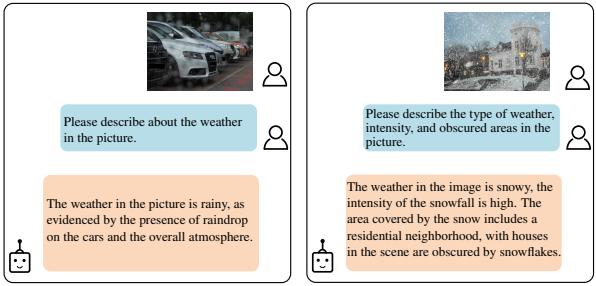

\begin{table*}[!h]
\centering
\caption{\small \it Quantitative comparison on the All-weather dataset. We respectively color the best and the second-best methods in \textcolor{red}{red} and \textcolor{blue}{blue}. }
\vspace{-1em}
\setlength{\tabcolsep}{8pt}
\small
\setlength{\tabcolsep}{12.8pt}
\begin{tabular}{c|c|cc|cc|cc|cc}
\Xhline{4\arrayrulewidth}
\multirow{2}{*}{Type} & \multirow{2}{*}{Method} & \multicolumn{2}{c|}{\textbf{Rain}} & \multicolumn{2}{c|}{\textbf{Snow}} & \multicolumn{2}{c|}{\textbf{Raindrop}} & \multicolumn{2}{c}{\textbf{Average}} \\
                   &                    &PSNR &SSIM &PSNR &SSIM &PSNR &SSIM &PSNR &SSIM               \\ \Xhline{4\arrayrulewidth}
                    \multicolumn{2}{c|}{BestT + VL}  &24.33   &0.860   &28.47   &0.872  &29.23   &0.895    &27.34 &0.876\\
                    \multicolumn{2}{c|}{BestT + GT}  &27.04   &0.913   &30.61   &0.900  &31.63   &0.936    &29.76 &0.916  \\ \Xhline{3\arrayrulewidth}
\multirow{5}{*}{\rotatebox{90}{General}}  &MPRNet \cite{zamir2021multi}      &23.08   &0.839   &27.69   &0.849  &28.75   &0.879    &26.51 &0.856               \\
                                &NAFNet \cite{chen2022simple}    &23.21   &0.840   &27.68   &0.847  &28.90   &0.890   &26.60 &0.859               \\
                                &Uformer \cite{wang2022uformer}    &22.93   &0.835   &27.50   &0.838  &28.51   &0.871   &26.31 &0.848               \\
                                &Restormer \cite{zamir2022restormer}  &23.37   &0.845   &27.81   &0.850  &29.10   &0.890    &26.76 &0.862               \\
                                &GRL \cite{li2023efficientgrl}        &23.31   &0.842   &27.79   &0.849  &29.05   &0.888    &26.72 &0.860               \\ \Xhline{3\arrayrulewidth}
\multirow{5}{*}{\rotatebox{90}{All-in-One}}     &All-in-One \cite{li2020all}    &24.71     &0.898     &28.33     &0.882     &31.12     &0.927     &28.05 &0.902  \\
                                &AirNet \cite{li2022all}    &23.12     &0.837     &27.92     &0.858     &28.23     &0.892     &26.42 &0.862        \\
                                &TUM \cite{chen2022learning}     & 23.92    &0.855     &29.27     &0.884     &30.75     &0.912     &27.98 &0.884        \\
                                &Transweather \cite{valanarasu2022transweather}     &23.18     &0.841     &27.80     &0.854    &28.98     &0.902    &26.65 &0.866       \\
                                &WGWS \cite{zhu2023learning}    &\textcolor{blue}{25.31}     &\textcolor{blue}{0.901}     &\textcolor{blue}{29.71}     &\textcolor{blue}{0.894}     &\textcolor{blue}{31.31}     &\textcolor{blue}{0.932}     &\textcolor{blue}{28.78} &\textcolor{blue}{0.909 }       \\
                                &Ours     &\textcolor{red}{26.92}     &\textcolor{red}{0.912 }    &\textcolor{red}{30.79}    &\textcolor{red}{0.905}     &\textcolor{red}{31.54}     &\textcolor{red}{0.933}     &\textcolor{red}{29.75} &\textcolor{red}{0.916}      \\ \Xhline{4\arrayrulewidth}
\end{tabular}
\label{tab:result_All-in-One}
\vspace{-0.5em}
\end{table*}

\subsection{Degradation Prior}
\paragraph{Degradation Prior Generation.}~We leverage the context learning capability of the PVL model (\eg, \cite{lai2023lisa})
to reason diverse degradation knowledge of the degraded image $\mathbf{I}^{\text{d}}$ 
with a question prompt $\mathbf{T}$.  
We format the question prompt $\mathbf{T}$ inspired by the chain-of-thought reasoning  that makes the model to 
identify the occurrence, type, and severity of degradation. 
In Fig.~\ref{img:vqa}, we provide the text description examples obtained by using different prompts. 
{The generated descriptions $\mathbf{P}^{\text{txt}}$ are defined as}
%
\begin{align}
    \mathbf{P}^{\text{txt}} &= \text{VL}(\mathbf{I}^{\text{c}}, \mathbf{T}) \ , \quad \mathbf{P}^{\text{txt}} \in \mathbb{R}^{L \times C^\text{vl}} \ ,
    \vspace{-2.5mm}
\end{align}
where VL$(\cdot, \cdot)$ is the PVL model, $L$ is the description length, and $C^\text{vl}$ is the channel dimension.
To preserve the representation {capabilities} of PVL, we prune out the text description output layer of PVL, and use the embedding before the output layer as $\mathbf{P}^{\text{txt}}$.


\vspace{-5mm}
\paragraph{Degradation Prior Embedding.} We align $\mathbf{P}^{\text{txt}}$ to the embedding space with size $C$ of our restoration model by using a multilayer perceptron network  $\text{MLP}(\cdot)$. We have
%
\begin{align}
    \mathbf{P}^{\text{emb}} = \text{MLP}(\mathbf{P}^{\text{txt}}) \ , \quad \mathbf{P}^{\text{emb}}  \in \mathbb{R}^{L \times C} \ .
\vspace{-3.5mm}
\end{align}
%
%
The $\mathbf{P}^{\text{emb}}$ is then used {as the degradation prior} in our restoration model to adaptively recover $\hat{\mathbf{I}}^{\text{c}}$.

\subsection{Language-driven Restoration Model}
We use an encoder-decoder architecture \cite{cui2022selective} as a backbone, and adaptively recover $\hat{\mathbf{I}}^{\text{c}}$ in the embedding space with $\mathbf{P}^{\text{emb}}$ and 
$\mathbf{X}$ (the encoded embedding of $\mathbf{I}^{\text{d}}$) as inputs. 

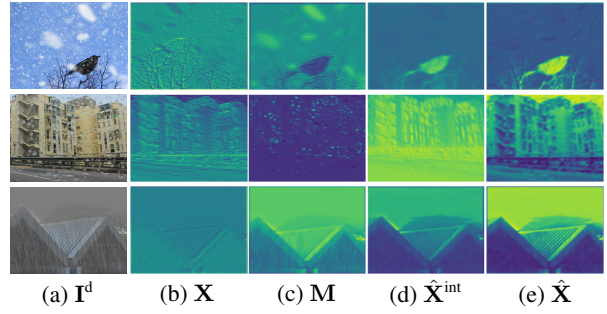
\begin{figure}[!t]
    \centering
    \begin{tikzpicture}
    \node[anchor=south west,inner sep=0] (image) at (0,0) {\includegraphics[width=0.95 \linewidth]{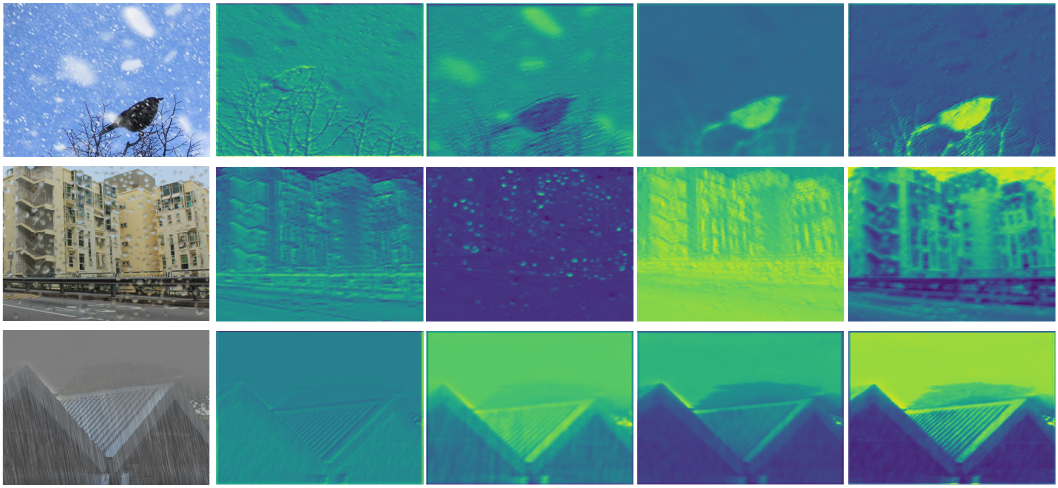}};
    \begin{scope}[x={(image.south east)},y={(image.north west)}] 
        \draw (0.10,  -0.063) node {\small (a) $\mathbf{I}^{\text{d}}$};
        \draw (0.30,  -0.07) node {\small (b) $\mathbf{X}$ };
        \draw (0.50,  -0.07) node {\small (c) $\mathbf{M}$ };
         \draw (0.70,  -0.0595) node {\small (d) $\hat{\mathbf{X}}^\text{int}$};
         \draw (0.90,  -0.0595) node {\small (e)  $\hat{\mathbf{X}}$};
    \end{scope}
    \end{tikzpicture}
    \vspace{-1em}
    \caption{\small \it
    Visualization of $\mathbf{X}$, $\mathbf{M}$, $\hat{\mathbf{X}}^{\text{int}}$ and $\hat{\mathbf{X}}$. Without cherry-picking, we show the channel with maximum gradients. From (b) $\mathbf{X}$ to (c) $\mathbf{M}$, degraded regions are highlighted, and regions with similar degradation severity have similar activation values on the feature map $\mathbf{M}$. From  (b) $\mathbf{X}$ to  (d) $\hat{\mathbf{X}}^\text{int}$ and  (e) $\hat{\mathbf{X}}$, degradations are  removed step by step. 
    }
    \label{img:mask}
    \vspace{-0.5em}
\end{figure}

\vspace{-5mm}
\paragraph{Degradation Map Measurement.}~The degradation prior $\mathbf{P}^{\text{emb}}$ is text description related with high-level semantics. However, for $\mathbf{X} \in \mathbb{R}^{H \times W \times C}$ with height $H$, width $W$, and channel $C$, the degradation is usually pixel-wise and distinct in severity. To allow pixel-wise computation of adverse weather removal, we measure the 2D degradation map $\mathbf{M}$ for $\mathbf{X}$ by the cross-attention mechanism, 
%
\begin{align}
    &\mathbf{Q} = \mathbf{X} \mathbf{W}^{\text{q}_1} \ , \quad \mathbf{K} = \mathbf{\mathbf{P}^{\text{emb}}} \mathbf{W}^{\text{k}_1} \ , \quad \mathbf{V} = \mathbf{\mathbf{P}^{\text{emb}}} \mathbf{W}^{\text{v}_1} \ , \\
    &\mathbf{M} = \text{Attention}(\mathbf{Q}, \mathbf{K}, \mathbf{V}) = \text{Softmax}\left(\mathbf{Q} \mathbf{K}^{\top} \right) \mathbf{V} \ ,   \label{eq:measurment}
\vspace{-2.5mm}
\end{align}
where $\mathbf{W}^{\text{q}_1}$, $\mathbf{W}^{\text{k}_1}$, and $\mathbf{W}^{\text{v}_1} \in \mathbb{R}^{C \times C}$ are linear projection matrices for obtaining the query, key, and value. 
Our cross-attention mechanism first  computes the semantic alignments between $\mathbf{X}$ and $\mathbf{P}^{\text{emb}}$ via $\text{Softmax}\left(\mathbf{Q} \mathbf{K}^{\top} \right)$,
{and then transforms $\mathbf{P}^{\text{emb}}$ into a 2D degradation map $\mathbf{M}$ that describes pixel-wise degradation of $\mathbf{X}$.}
We visualize $\mathbf{X}$ and $\mathbf{M}$ in Fig.~\ref{img:mask}. The regions of similar degradation severity have been individually highlighted, by applying Eq.~\eqref{eq:measurment}. 

\vspace{-5mm}
\paragraph{Top-K Expert Restoration.}~{Not all pixels degrade equally, and we adaptively restore each of them by using pixel-wise degradation knowledge within $\mathbf{M}$.}
We have $N$ candidate experts (convolution filters), $\{\mathbf{F}_{n}|n=1,...,N\}$, for diverse adverse weather conditions, where each candidate $\mathbf{F}_{n}$ is expertise at generating restoration features for specific degradation types or severities (see \ref{sec:ablation} ). 
With the degradation map $\mathbf{M}$, we select experts for each pixel of $\mathbf{X}$. 

Specifically, the degradation map $\mathbf{M} \in \mathbb{R}^{H \times W \times C}$ is fed to a feedforward network $\text{FFN}^{\text{m}}$, followed by a Softmax layer along the last dimension, to output a normalized pixel-wise selection score $\mathbf{S} \in \mathbb{R}^{H \times W \times N}$, \ie, $\mathbf{S} = \text{Softmax}(\text{FFN}^{\text{m}}(\mathbf{M}))$. For a pixel with location $(i,j), i\in[1,H], j\in[1,W]$, $\mathbf{S}(i,j) \in \mathbb{R}^N$ and $\mathbf{S}(i,j,n)$ measures how likely the $n$-th expert $\mathbf{F}_{n}$ can be used to describe the degradation of $\mathbf{X}(i,j)$.
%
%

To find the best experts describing the degradation of $\mathbf{X}(i,j)$, we compute the Top-K scores of $\mathbf{S}(i,j)$ and record corresponding indices as the best experts. The best experts restored feature for location $(i,j)$ is given by,
\begin{align}
    &\hat{\mathbf{X}}^{\text{int}}(i,j) = \sum_{k=1}^{K} \mathbf{S}(i,j,\rho(k)) \cdot \mathcal{E}(i, j, \rho(k)) \ , \\
    &\mathcal{E}(i, j, \rho(k)) = \sum_{\Delta i, \Delta j} \mathbf{X}(i + \Delta i, j + \Delta j) \cdot  \mathbf{F}_{\rho(k)}(\Delta i, \Delta j) \ , \nonumber
\end{align}
where $\rho(k)$ is the index of the selected $k$-th  expert.  $\mathcal{E}(i, j, \rho(k))$ is the  convolution result of $\mathbf{X}(i,j)$ and the $k$-th expert $\mathbf{F}_{\rho(k)}$, and $(\Delta i, \Delta j)$ iterates over the convolution kernel size. $\mathbf{S}(i,j,\rho(k))$ are weights for prioritizing different experts. With our pixel-wise Top-K experts, the intermediate restoration feature $\hat{\mathbf{X}}^{\text{int}}$ for each pixel of $\mathbf{X}$ is adaptively generated. 



\vspace{-5mm}
\paragraph{Restoration Feature Aggregation.} Pixels with similar degradation {information} in $\mathbf{M}$ 
potentially benefit each other in deriving the restoration $\hat{\mathbf{X}}$. 
We compute the compatibility between degradation prior  $\mathbf{M}$ and restoration features $\hat{\mathbf{X}}^{\text{int}}$, and aggregate weighted restoration features pixel-wisely by using the cross-attention mechanism, 
\begin{align}
    &\mathbf{Q} = \mathbf{M} \mathbf{W}^{\text{q}_2} \ , \quad \mathbf{K} = \mathbf{\hat{\mathbf{X}}^{\text{int}}} \mathbf{W}^{\text{k}_2} \ , \quad \mathbf{V} = \mathbf{\hat{\mathbf{X}}^{\text{int}}} \mathbf{W}^{\text{v}_2} \ , \\
    &\hat{\mathbf{X}} = \text{FFN}^\text{int}(\text{Attention}(\mathbf{Q}, \mathbf{K}, \mathbf{V})) \ ,  
\end{align}
where $\text{FFN}^\text{int}$ is a feedforward convolutional network to improve the feature locality. Finally, $\hat{\mathbf{X}}$ is decoded to the restoration $\hat{\mathbf{I}}^{\text{c}}$. In Fig.~\ref{img:mask}, we compare  $\hat{\mathbf{X}}^{\text{int}}$ and $\hat{\mathbf{X}}$, and  find that $\hat{\mathbf{X}}$ is cleaner than $\hat{\mathbf{X}}^{\text{int}}$, as pixels with similar degradation measurements benefit each other in the restorations.

\subsection{Loss Function}
We train our network with Charbonnier loss ${\mathcal{L}}_{\text{char}}$ and gradient-level edge loss ${\mathcal{L}}_{\text{edge}}$, to penalize the deviation of restoration from ground truth clean image, and encourage the consistent image gradients. We have ${\mathcal{L}}_{\text{char}}$ as 
%
\begin{equation}
\mathcal{L}_{\text{char}}= \sqrt{\lVert \mathbf{I}^{\text{c}} - {\mathbf{\hat{I}}}^{\text{c}} \rVert^{2}+ \varepsilon^{2}} \ , 
\vspace{-2.5mm}
\end{equation}
where $\varepsilon = 10^{-4}$ is a constant in all experiments. 
The ${\mathcal{L}}_{\text{edge}}$ is
%
\begin{equation}
\mathcal{L}_{\text{edge}}= \sqrt{\lVert \nabla \mathbf{I}^{\text{c}} - \nabla {\mathbf{\hat{I}}}^{\text{c}}  \rVert^{2}+ \varepsilon^{2}} \ ,
\end{equation}
where $\nabla$ is the laplacian gradient operator. With a balance parameter $\lambda$, the total loss is given by
%
\begin{equation}
\mathcal{L}_{\text{total}}= \mathcal{L}_{\text{char}}+ \lambda{\mathcal{L}_{\text{edge}}} \ . 
\vspace{-2.5mm}
\end{equation}
%

\begin{table*}[!h]
\centering
\caption{\small \it Quantitative comparison on the WeatherStream dataset. We color the best and the second-best methods in \textcolor{red}{red} and \textcolor{blue}{blue}.}
\vspace{-1.0em}
\setlength{\tabcolsep}{8pt}
\small
\setlength{\tabcolsep}{12pt}
\begin{tabular}{c|c|cc|cc|cc|cc}
\Xhline{4\arrayrulewidth}
\multirow{2}{*}{Type} & \multirow{2}{*}{Method} & \multicolumn{2}{c|}{\textbf{Rain}} & \multicolumn{2}{c|}{\textbf{Haze}} & \multicolumn{2}{c|}{\textbf{Snow}} & \multicolumn{2}{c}{\textbf{Average}} \\
                   &                    &PSNR &SSIM &PSNR &SSIM &PSNR &SSIM &PSNR &SSIM               \\ \Xhline{4\arrayrulewidth}
                    \multicolumn{2}{c|}{BestT + VL}  &21.20 &0.781  &21.60 &0.755  &20.32 &0.772  &21.04 &0.769\\
                    \multicolumn{2}{c|}{BestT + GT}  &23.95 &0.810  &22.97 &0.804  &22.70 &0.828  &23.21 &0.814  \\ \Xhline{3\arrayrulewidth}
\multirow{5}{*}{\rotatebox{90}{General}}  &MPRNet \cite{zamir2021multi}     &21.50 &0.791 &21.73 &0.763 &20.74 &0.801     &21.32 &0.785               \\
                                &NAFNet \cite{chen2022simple}     &23.01 &0.803 &22.20 &0.803 &22.11 &0.826     &22.44 &0.811               \\
                                &Uformer \cite{wang2022uformer}    &22.25 &0.791 &18.81 &0.763 &20.94 &0.801 &20.67 &0.785              \\
                                &Restormer \cite{zamir2022restormer}     &23.67 &0.804 &22.90 &0.803 &22.51 &0.828    &22.86 &0.812              \\
                                &GRL \cite{li2023efficientgrl}     &23.75 &0.805 &22.88 &0.802 &22.59 &0.829    &23.07 &0.812               \\ \Xhline{3\arrayrulewidth}
\multirow{5}{*}{\rotatebox{90}{All-in-One}}     &All-in-One \cite{li2020all}     &-     &-     &-     &-     &-     &-     &-    &-   \\
                                &AirNet \cite{li2022all}    &22.52 &0.797 &21.56 &0.770 &21.44 &0.812     &21.84 &0.793       \\
                                &TUM \cite{chen2022learning}     &23.22 &0.795 &22.38 &0.805 &22.25 &0.827     &22.62 &0.809       \\
                                &Transweather \cite{valanarasu2022transweather}    &22.21 &0.772 &22.55 &0.774 &21.79 &0.792  &22.18 &0.779      \\
                                &WGWS \cite{zhu2023learning}    &\textcolor{blue}{23.80} &\textcolor{blue}{0.807} &\textcolor{blue}{22.78} &\textcolor{blue}{0.800} &\textcolor{blue}{22.72} &\textcolor{blue}{0.831}    &\textcolor{blue}{23.10} &\textcolor{blue}{0.813}       \\
                                &Ours     &\textcolor{red}{24.42} &\textcolor{red}{0.818} &\textcolor{red}{23.11} &\textcolor{red}{0.809} &\textcolor{red}{23.12}  &\textcolor{red}{0.838}     &\textcolor{red}{23.55}&\textcolor{red}{0.822}    \\ \Xhline{4\arrayrulewidth}
\end{tabular}
\label{tab:result_WeatherStream}
\vspace{-1.11em}
\end{table*}

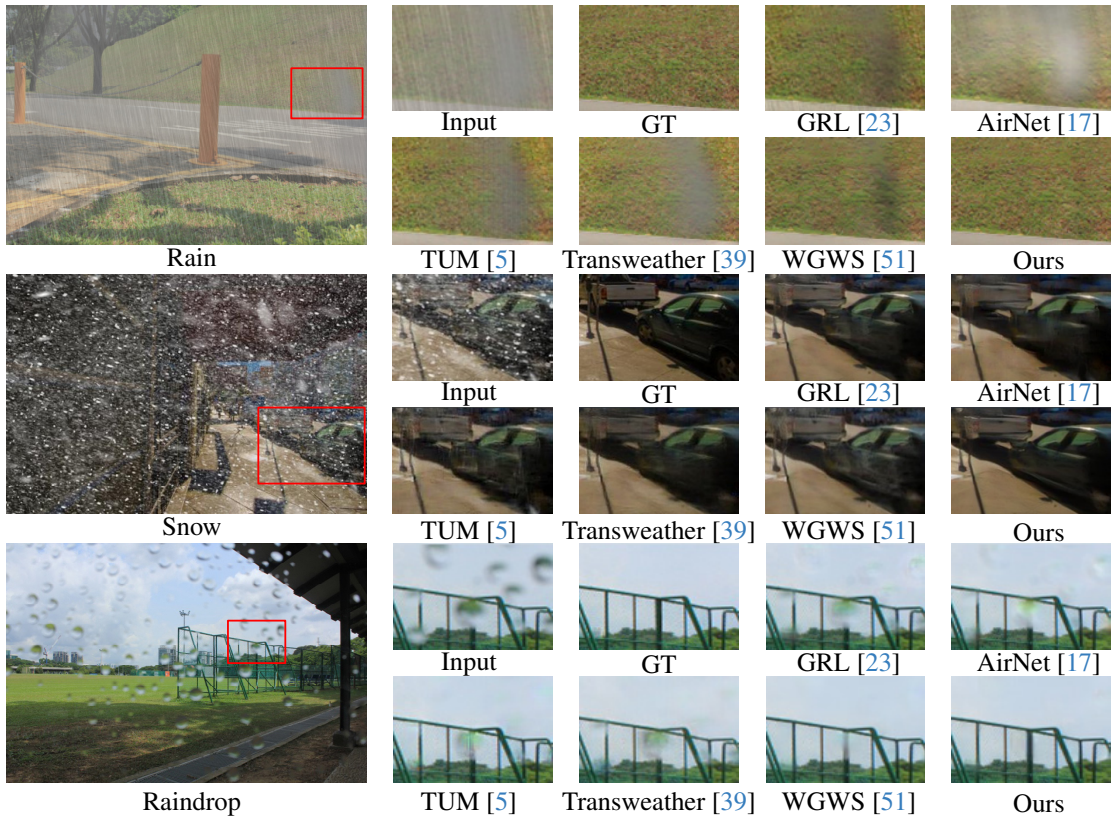
\begin{figure*}[!t]
    \centering
    \begin{tikzpicture}
    \node[anchor=south west,inner sep=0] (image) at (0,0) {\includegraphics[width=0.85 \linewidth]{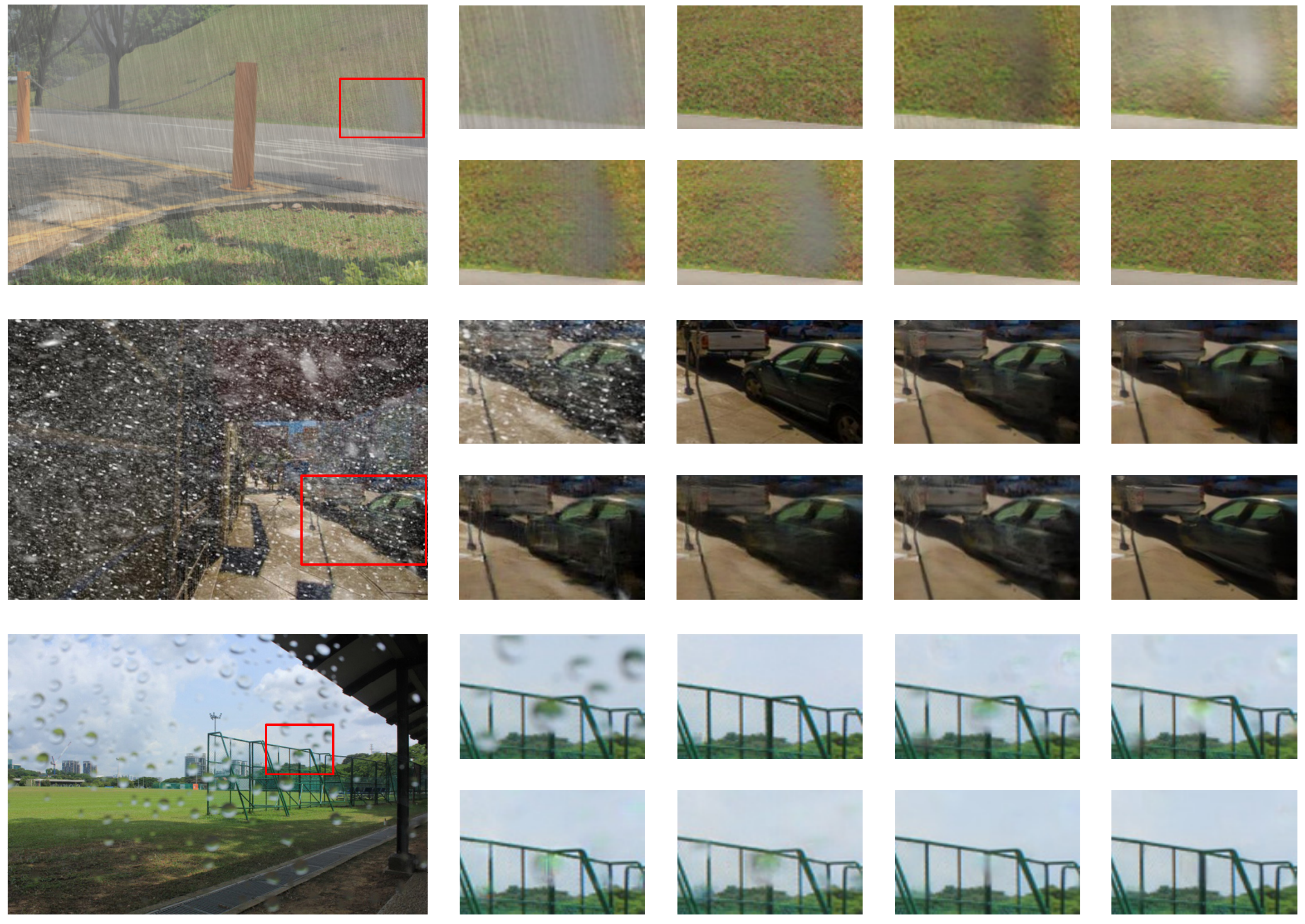}};
    \begin{scope}[x={(image.south east)},y={(image.north west)}] 
        \draw (0.172,  -0.016) node {Raindrop};
        \draw (0.172,  0.335) node {Snow};
        \draw (0.172,  0.675) node {Rain};

    \foreach \r in {0,...,2}
        \foreach \mylabel [count=\x from 0] in {TUM \cite{chen2022learning}, Transweather \cite{valanarasu2022transweather}, WGWS \cite{zhu2023learning}, Ours}
            \draw (0.42 + \x * 0.17, \r * 0.343 -0.015) node {\mylabel};

    \foreach \r in {0,...,2}
        \foreach \mylabel [count=\x from 0] in {Input, GT, GRL \cite{li2023efficientgrl}, AirNet \cite{li2022all}}
            \draw (0.42 + \x * 0.17, \r * 0.341 +0.161) node {\mylabel};

    \end{scope}
    \end{tikzpicture}
    \vspace{-1em}
    \caption{\small \it 
    Qualitative comparison on the All-weather dataset. The first column shows degraded images, while the crops for the bounding box regions of degraded images, ground truth, restoration from SOTA  methods and our method are shown in the subsequent columns.
    }
    \label{img:compareAll-in-One}
    \vspace{-0.5em}
\end{figure*}

\section{Experiments and Analysis}

\paragraph{Implementation Details.}
Our model is implemented using the PyTorch framework and all experiments are conducted on an RTX A6000 GPU. We train our network with batch size $4$ and $4 \times 10^{6}$ iterations using the ADAM optimizer. The learning rate is decayed from $2 \times 10^{-4}$ to $1 \times 10^{-6}$ by cosine annealing strategy. In training, images are randomly cropped to size $256 \times 256$, and $\lambda = 0.05$.

\vspace{-5mm}
\paragraph{Datasets.} 
Our experiments are conducted on both synthetic dataset and real dataset, \ie, All-weather dataset \cite{li2020all} and WeatherStream \cite{zhang2023weatherstream} dataset. All-weather dataset {comprises} $18,609$ training images and $17,609$ testing images, and is composed of subsets from Outdoor-rain \cite{li2019heavy}, Snow100K-L \cite{liu2018desnownet}, and Raindrop \cite{qian2018attentive}, corresponding to rain, snow, and raindrops weather conditions respectively. WeatherStream dataset is a real-world dataset that has three weather conditions, \ie, rain, snow, and fog, with a total of $176,100$ training images and $11,400$ testing images.

\vspace{-5mm}
\paragraph{Evaluation Metrics.}
We use an average of peak signal-to-noise ratio (PSNR) and structural similarity (SSIM) as our evaluation metrics. The higher, the better.

\vspace{-5mm}
\paragraph{Baseline Methods.} We compare with the state-of-the-art (SOTA) general and all-in-one methods. For general methods, task-specific methods are trained with multi-task learning of different weather conditions, \ie, MPRNet \cite{zamir2021multi}, NAFNet \cite{chen2022simple}, Uformer\cite{wang2022uformer}, Restormer \cite{zamir2022restormer}, and GRL \cite{li2023efficientgrl}. All-in-one methods are All-in-One \cite{li2020all}, AirNet\cite{li2022all}, TUM \cite{chen2022learning}, Transweather \cite{valanarasu2022transweather}, and WGWS \cite{zhu2023learning}.

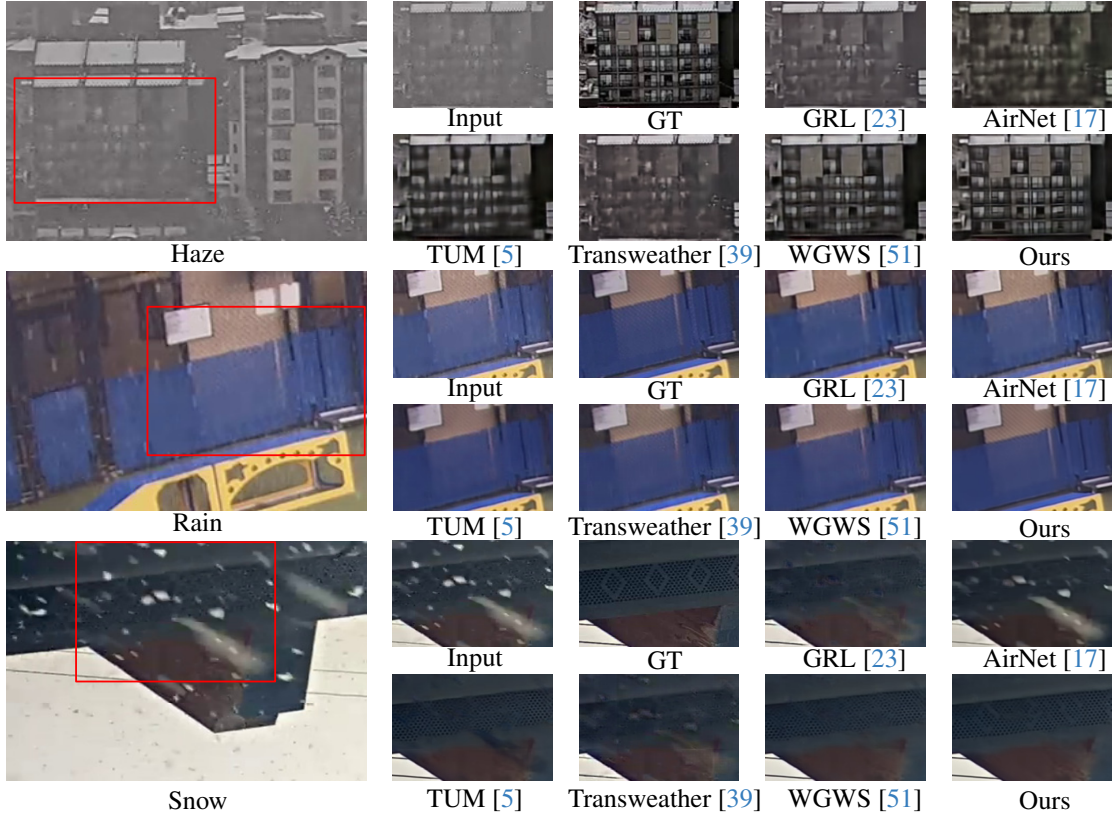
\begin{figure*}[!t]
    \centering
    \begin{tikzpicture}
    \node[anchor=south west,inner sep=0] (image) at (0,0) {\includegraphics[width=0.85 \linewidth]{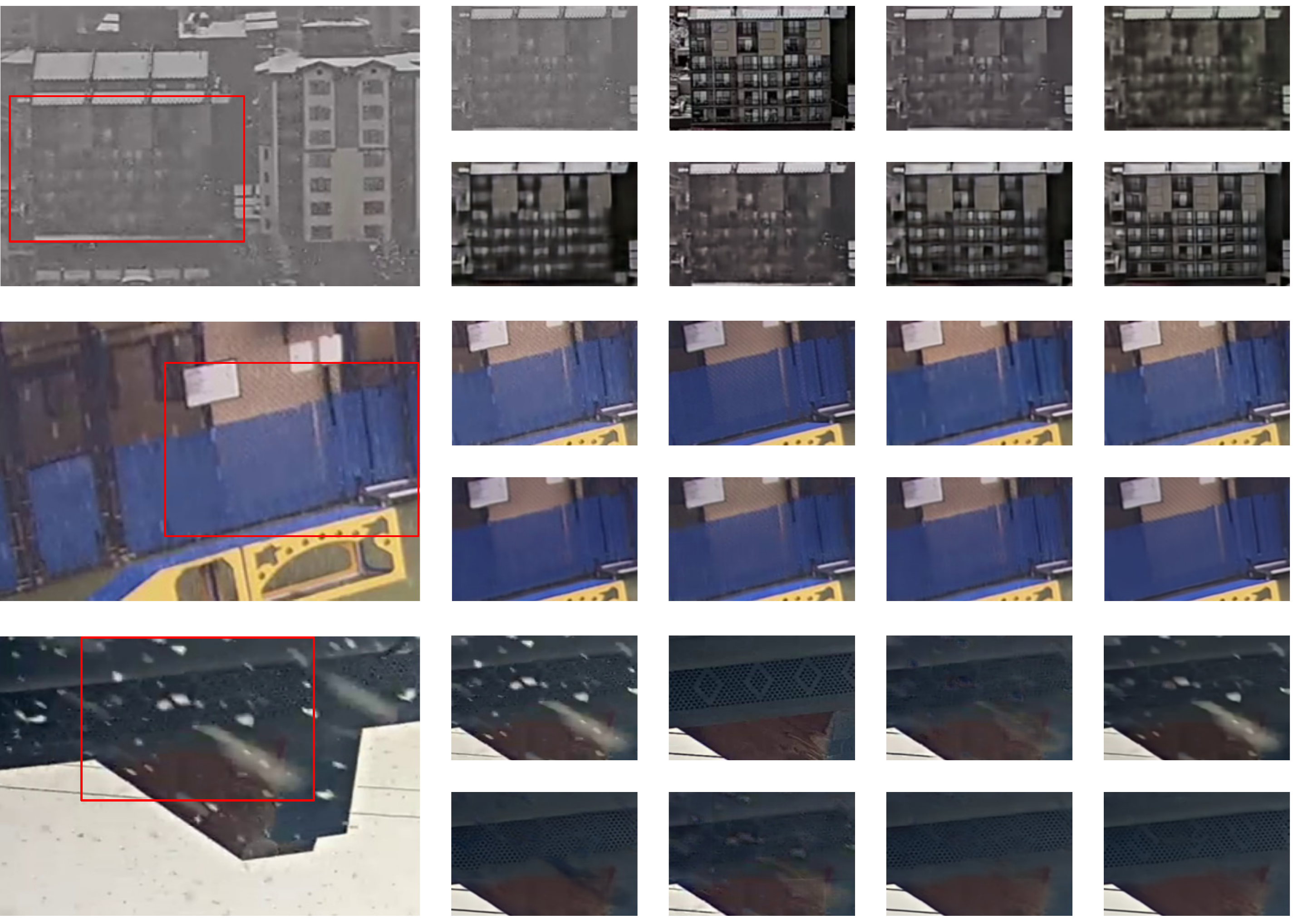}};
    \begin{scope}[x={(image.south east)},y={(image.north west)}] 
        \draw (0.172,  -0.016) node {Snow};
        \draw (0.172,  0.335) node {Rain};
        \draw (0.172,  0.675) node {Haze};
    \foreach \r in {0,...,2}
        \foreach \mylabel [count=\x from 0] in {TUM \cite{chen2022learning}, Transweather \cite{valanarasu2022transweather}, WGWS \cite{zhu2023learning}, Ours}
            \draw (0.42 + \x * 0.17, \r * 0.343 -0.015) node {\mylabel};

    \foreach \r in {0,...,2}
        \foreach \mylabel [count=\x from 0] in {Input, GT, GRL \cite{li2023efficientgrl}, AirNet \cite{li2022all}}
            \draw (0.42 + \x * 0.17, \r * 0.341 +0.161) node {\mylabel};

    \end{scope}
    \end{tikzpicture}
    \vspace{-1em}
    \caption{\small \it 
    Qualitative comparison on the WeatherStream dataset. The first column shows degraded images, while the crops for the bounding box regions of degraded images, ground truth, restoration from SOTA  methods and our method are shown in the subsequent columns.
    }
    \label{img:compareWeatherStream}
    \vspace{-0.5em}
\end{figure*}

\subsection{Experimental Results}

\paragraph{Quantitative Comparison.}  
We compare with the SOTA general and all-in-one methods on the All-weather and WeatherStream datasets in Tab.~\ref{tab:result_All-in-One} and Tab.~\ref{tab:result_WeatherStream}, respectively. Our method achieves the best performance.

Furthermore, we create two strong baselines:
i) BestT + VL. We use a PVL model for zero-shot adverse weather classification, and select the best task-specific methods for restoring degraded images;
ii) BestT + GT. The ground truth adverse weather type is used to select the best task-specific method for image restoration.  
We find that 
i) Our method significantly outperforms `BestT + VL', though it uses the same PVL as our model and uses multiple models for different adverse weather conditions. This shows that non-trivial designs are required for effectively leveraging the PVL model in all-in-one adverse weather removal;
ii) Although `BestT + GT' outperforms most SOTA general and all-in-one methods, {our method} still achieves competitive performance compared to it. This indicates:  1) the benefit of accessing prior knowledge of adverse weather conditions, and 2) our method has learned the shared and weather-specific knowledge adaptively by selecting sparse experts dynamically.

\vspace{-5mm}
\paragraph{Qualitative Comparison.} We compare with the SOTA methods on restoring images degraded by different adverse weather conditions. The results are given  in Fig.~\ref{img:compareAll-in-One} and Fig.~\ref{img:compareWeatherStream}. Our method consistently restores clearer images than SOTA methods under different weather conditions.

\subsection{Ablation studies and Discussions}
\label{sec:ablation}
We validate the effectiveness and components of our framework on the All-weather dataset. {{Due to} space limitations, the evaluations for PVL methods, prompt questions, expert designs, and mixed degradations are given in the  supplementary material.}

\begin{table}[!t]
  \centering
  \caption{\small \it The effectiveness of our model components.}
  \vspace{-1em}
   \small 
   \setlength{\tabcolsep}{13.1pt}
    \begin{tabular}{c|c|c|c|c}
\Xhline{4\arrayrulewidth}
     DMM &  TER & RFA &PSNR &SSIM   \\
\Xhline{3\arrayrulewidth}
    \xmark  & \cmark  & \xmark    & 27.93 & 0.882\\
    \xmark  & \xmark  &\cmark     & 28.37 & 0.889  \\
    \cmark  & \cmark  & \xmark    & 28.25 & 0.890 \\
    \cmark  & \xmark  &\cmark     &\textcolor{blue}{29.11}  &\textcolor{blue}{0.902}  \\
    \xmark  & \cmark  &\cmark     & 28.55 & 0.895\\
    \cmark  &\cmark   &\cmark     &\textcolor{red}{29.75}  &\textcolor{red}{0.916}  \\
\Xhline{4\arrayrulewidth}
    \end{tabular}%
  \label{tab:ab1}%
\end{table}%

\begin{figure*}[!t]
    \centering
    \includegraphics[width=\linewidth]{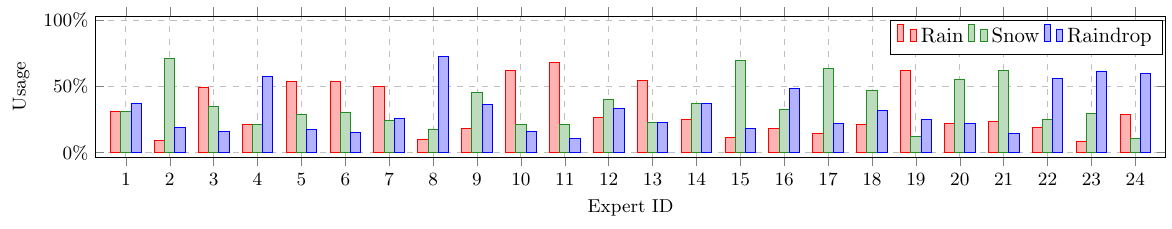}
    \vspace{-2.3em}
    \caption{\small \it The usage frequencies of experts for different degradation types. For example, Expert 8, 11, and 15 focuses on raindrop, rain, and snow, respectively. Expert 12 handles all types of degradation. Please refer to Fig.~\ref{img:experts} for visualization. }
    \label{fig:usage}
    \vspace{-0.8em}
\end{figure*}

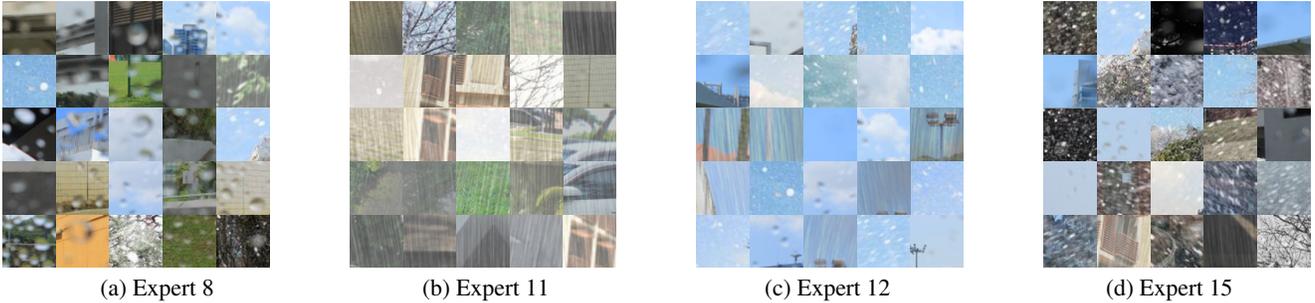
\begin{figure*}[!t]
    \centering
    \vspace{-0.5em}
    \begin{tikzpicture}
    \node[anchor=south west,inner sep=0] (image) at (0,0) {\includegraphics[width= \linewidth]{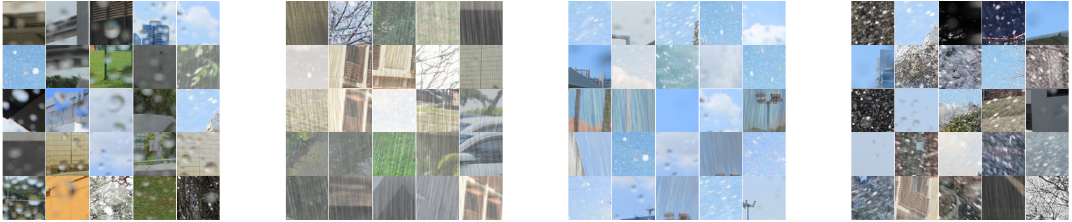}};
    \begin{scope}[x={(image.south east)},y={(image.north west)}] 
        \draw (0.12,  -0.07) node {\small (a) Expert 8};
        \draw (0.37,  -0.07) node {\small (b) Expert 11};
        \draw (0.63,  -0.07) node {\small (c) Expert 12};
         \draw (0.89,  -0.07) node {\small (d) Expert 15};

    \end{scope}
    \end{tikzpicture}
    \vspace{-2.0em}
    \caption{\small \it Sample image regions, activating different experts. We show regions with highest selection scores for experts 8, 11, 12, and 15 on the All-weather dataset. 
    }
    \label{img:experts}
    \vspace{-0.5em}
\end{figure*}

\vspace{-5mm}
\paragraph{Model Architecture.} We study the effectiveness of the degradation map measurement (DMM), top-K expert restoration (TER), and restoration feature aggregation (RFA) modules. The results are given in Tab.~\ref{tab:ab1}. The best performance is achieved by using all modules.

\begin{table}[!t]
\centering
\caption{\small \it Comparisons on the All-weather dataset with degradation severity of slight, moderate, and heavy.}
\vspace{-1em}
\setlength{\tabcolsep}{4pt}
\small
\setlength{\tabcolsep}{5.5pt}
\begin{tabular}{c|cc|cc|cc}
\Xhline{4\arrayrulewidth}
\multirow{2}{*}{Method} & \multicolumn{2}{c|}{\textbf{Slight}} & \multicolumn{2}{c|}{\textbf{Moderate}} & \multicolumn{2}{c}{\textbf{Heavy}} \\
                                     &PSNR &SSIM &PSNR &SSIM &PSNR &SSIM                \\ \Xhline{4\arrayrulewidth}
AirNet      &27.59	&0.895	&26.49	&0.865	&24.50	&0.818  \\  
WGWS        &\textcolor{blue}{29.74}	&\textcolor{blue}{0.921}	&\textcolor{blue}{28.77}	&\textcolor{blue}{0.910}	&\textcolor{blue}{27.14}	&\textcolor{blue}{0.886}  \\  
Ours      &\textcolor{red}{30.46}	&\textcolor{red}{0.925}	&\textcolor{red}{29.78}	&\textcolor{red}{0.918}	&\textcolor{red}{28.38}	&\textcolor{red}{0.899}   \\  
\Xhline{4\arrayrulewidth}
\end{tabular}
\label{tab:Severity}
\vspace{-0.5em}
\end{table}

\vspace{-5mm}
\paragraph{Degradation Severity.} 
We use the PVL model to partition the All-weather dataset \cite{li2020all} into three subsets based on degradation severity: slight, moderate, and heavy. We then compare our results with those of AirNet and WGWS in Tab.~\ref{tab:Severity}.
Our method shows significant improvement over AirNet \cite{li2022all} and WGWS \cite{zhu2023learning} in the heavily degraded subset. This indicates the effectiveness of reasoning with diverse weather-specific knowledge, such as severity. 
   
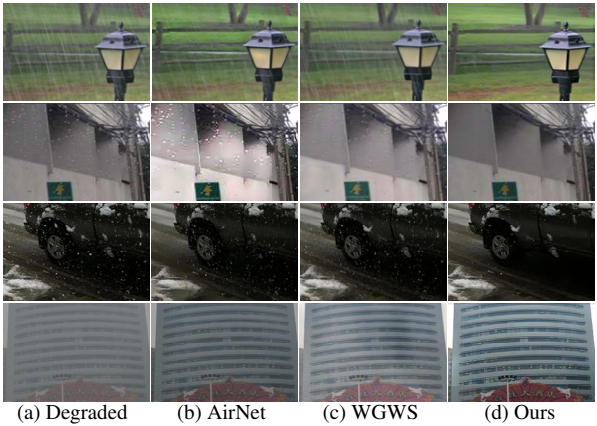
\begin{figure}[!t]
    \centering
    \begin{tikzpicture}
    \node[anchor=south west,inner sep=0] (image) at (0,0) {\includegraphics[width=0.95 \linewidth]{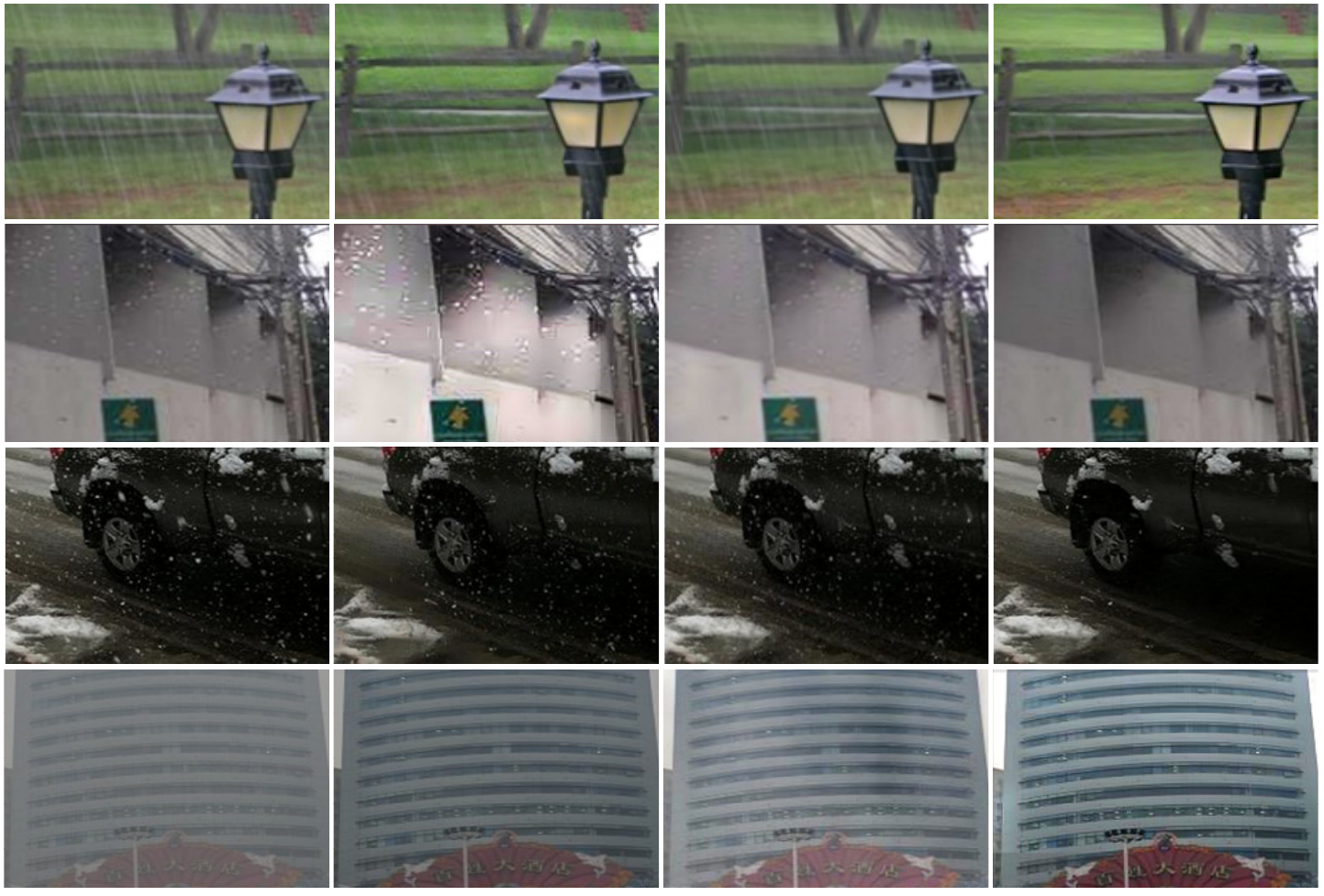}};
    \begin{scope}[x={(image.south east)},y={(image.north west)}] 
        \draw (0.12,  -0.02) node {\footnotesize (a) Degraded};
        \draw (0.37,  -0.02) node {\footnotesize (b) AirNet};
        \draw (0.62,  -0.02) node {\footnotesize (c) WGWS};
         \draw (0.87,  -0.02) node {\footnotesize (d) Ours};
    \end{scope}
    \end{tikzpicture}
    \vspace{-1em}
    \caption{\small \it From the first to last rows, the degradation is rain, raindrop, snow, and haze, respectively. From left to right, (a) degraded images are restored by (b) AirNet, (c) WGWS, and (d) our method.
    }
    \label{img:realrain}
    \vspace{-0.5em}
\end{figure}

\vspace{-5.5mm}
\paragraph{Pixel-wise Expert.} Different regions of the same degraded image often exhibit varying degrees of degradation severity, and should be adaptively and pixel-wisely restored. We use the degradation prior provided by the PVL  model to select the same expert for the degraded image. It shows a 0.33 dB/0.07 decrease in PSNR/SSIM, indicating the necessity of selecting experts pixel-wisely.

\vspace{-5mm}
\paragraph{Adaptive Adverse Weather Removal.} 
We study the model response to different adverse weather degradation by measuring the usage of our experts. Fig.~\ref{fig:usage} shows the usage for rain, snow, and raindrop. We find that there are experts for weather-specific knowledge and shared knowledge among degradation, such as Expert 8, Expert 11, Expert 12, and Expert 15. We visualize the image patches selecting the four experts in Fig.~\ref{img:experts}, where Expert 8, Expert 11, and Expert 15 are selected by image regions with raindrop, rain, and snow degradation, and Expert 12 is selected by image regions with sky in all types of degradation.


\vspace{-5.5mm}
\paragraph{Model Ability Under Complex Weather Condition.} 
We test our model trained on the All-weather dataset with real images degraded by rain, raindrop, snow, and haze, as shown in Fig.~\ref{img:realrain}. We compare our results with the two most competitive methods, AirNet and WGWS. Our model successfully disentangles weather-specific knowledge and generalizes to restore images degraded by haze.

\section{Conclusion and Broader Impact}

We have proposed an LDR framework that adaptively removes various adverse weather conditions in an all-in-one solution. Our key insight is to leverage a pre-trained vision-language model to reason diverse weather-specific knowledge in a degraded image. We then use this knowledge to restore a clean image with three modules: degradation map measurement, Top-K expert restoration, and restoration feature aggregation. Experiments on standard benchmark datasets demonstrate that our method outperforms past works by a large margin.



\vspace{-5.5mm}
\paragraph{Broader Impact.} Our method is promising to be developed as an image restoration foundation model, prompting by degradation prior generated by a vision-language model.




{
    \small
    \bibliographystyle{ieeenat_fullname}
    \bibliography{main}
}

\end{document}